\documentclass{article}

    \PassOptionsToPackage{numbers, compress}{natbib}

\usepackage[final]{neurips_2019}

\usepackage[utf8]{inputenc} 
\usepackage[T1]{fontenc}    
\usepackage{hyperref}       
\usepackage{url}            
\usepackage{booktabs}       
\usepackage{amsfonts}       
\usepackage{nicefrac}       
\usepackage{microtype}      

\usepackage{bbm}
\usepackage{mwe}
\usepackage{soul}
\usepackage{subcaption}
\usepackage{cleveref}
\usepackage{graphicx}
\usepackage[english]{babel}
\usepackage{color}
\usepackage{multirow}
\usepackage{algorithm}
\usepackage[noend]{algorithmic}

\usepackage{mathtools}

\usepackage{wrapfig}

\usepackage[
  separate-uncertainty = true,
  multi-part-units = repeat
]{siunitx}

\title{Post-Training 4-bit Quantization on Embedding Tables}

\author{%
  Hui Guan$^1$, Andrey Malevich$^2$, Jiyan Yang$^2$, Jongsoo Park$^2$, Hector Yuen$^2$ \\
  $^1$North Carolina State University\\
  $^2$Facebook, Inc \\
  \texttt{hguan2@ncsu.edu, \{amalevich, chocjy, jongsoo, hyz\}@fb.com} \\
}

\begin{document}

\maketitle

\begin{abstract}
Continuous representations have been widely adopted in recommender systems where a large number of entities are represented using embedding vectors. As the cardinality of the entities increases, the embedding components can easily contain millions of parameters and become the bottleneck in both storage and inference due to large memory consumption. This work focuses on post-training 4-bit quantization on the continuous embeddings. We propose row-wise uniform quantization with greedy search and codebook-based quantization that consistently outperforms state-of-the-art quantization approaches on reducing accuracy degradation. We deploy our uniform quantization technique on a production model in Facebook and demonstrate that it can reduce the model size to only 13.89\% of the single-precision version while the model quality stays neutral. 
  
\end{abstract}

\section{Introduction}


The success of word embeddings in Natural Language Processing (NLP)~\cite{pennington2014glove, mikolov2013distributed} has promoted the wide adoption of continuous representations in recommender systems in recent years. Embedding-based approaches have achieved state-of-the-art performance in recommendation and ranking tasks and have been successfully applied in real-world applications~\cite{naumov2019deep, park2018deep, cheng2016wide, lian2018xdeepfm, he2017neural}. 
In these recommendation models, a large number of entities such as page ids and user ids are encoded using \textit{embedding tables} whose \textit{row vectors} correspond to entities. As embedding tables scale with the number of entities and embedding dimensions, 
they can easily contain billions of parameters and usually contribute to 99.99\% of the size of the models. 
For example, a single-precision embedding table with 50,000,000 number of ids and 64 embedding dimensions costs 12GB memory and a recommendation model could contain up to hundreds of embedding tables. Furthermore, due to memory-bandwidth limitations, embedding table lookups are one of the most time-consuming operations. Their proportion will increase due to the acceleration of other parts (e.g. FC) from faster increase of compute throughput than memory bandwidth, posing a great challenge to get real-time predictions~\cite{park2018deep, naumov2019deep}.

Quantization is one of the effective approaches to reduce model size. By quantizing floating-point values in embedding tables to low-precision numbers that use less number of bits, a large recommendation model can be reduced to a model with much smaller model size and memory bandwidth consumption during inference. Prior work on quantization has been focusing on quantization-aware training from scratch or a pre-trained floating-point model~\cite{courbariaux1602binarynet, zhou2016dorefa,zhang2018training, he2016effective, micikevicius2017mixed, de2018high, li2017training, goncharenko2018fast}. Although these techniques have shown promising results, they are not always applicable in many practical scenarios where the training dataset might be no longer available during model deployment~\cite{zhao2019improving, banner2018aciq, choukroun2019low, migacz20178}. In these cases, post-training quantization is a more desirable approach. 
Post-training quantization is simple to use and convenient for rapid deployment. Recent studies have shown post-training quantization using 8-bit precision can achieve accuracy close to that of single-precision models in a wide variety of DNN architectures~\cite{migacz20178, krishnamoorthi2018quantizing}. Post-training quantization using lower bit width (e.g. 4-bit), however, usually incurs significant accuracy drop~\cite{choukroun2019low}.

Several state-of-the-art post-training quantization techniques that rely on the clipping have been proposed to mitigate accuracy degradation. Shin et al.~\cite{shin2016fixed} and Sung et al.~\cite{sung2015resiliency} approximated the inputs as a histogram and adopt a clipping threshold that minimizes the $\ell_2$ norm of the quantization error. Migacz et al.~\cite{migacz20178} proposed an iterative approach to search for the clipping threshold based on Kullback-Leibler Divergence measure for quantizing activations. Later, Banner et al.~\cite{banner2018aciq} proposed ACIQ, an analytic solution that computes the optimal clip threshold by assuming the input values are sampled from a Gaussian or Laplacian distribution. Although these approaches are demonstrated to reduce the accuracy drops to some extent, the problem of post-training 4-bit quantization without accuracy drop is still unsolved yet. Empirically, we also observe that the above-mentioned approaches can result in significant accuracy drops when applied to embedding table quantization.

In this paper, we explore a variety of post-training 4-bit quantization methods on embedding tables and propose novel quantization approaches that can reduce model size while incurring negligible accuracy degradation. Quantization on embedding tables is usually applied to row vectors (\textit{row-wise quantization}) to reduce quantization error.  Throughout the paper, quantization is applied to row vectors unless noted differently. We notice that the prior post-training quantization approaches approximate the inputs to quantize using either a histogram or some distributions. While these assumptions are beneficial to derive efficient algorithms for finding the optimal clipping thresholds for weights and activations of convolutional neural networks (CNNs), they are not suitable for embedding tables 
because their row vectors contain too few values to be well-characterized using either histograms or distributions. Inspired by these understandings, we design quantization algorithms that directly target at minimizing the mean square error after quantization. Specifically, 
\begin{itemize}
\item Our exploration reveals that state-of-the-art post-training 4-bit quantization approaches are no better than the approach that uses the range the input without clipping when the input contains only tens or hundreds of values, as in the case of embedding table quantization. 

\item We propose two simple yet effective approaches to improve 4-bit quantization on embedding tables: 1) row-wise uniform quantization with greedy search that finds the best clipping thresholds from a gradually discovered set of local optima; 2) codebook-based quantization that maps inputs to indices of non-uniformly distributed values using k-means clustering. 

\item Moreover, for 4-bit quantized embedding tables using uniform quantization, we can achieve dequantization performance similar to the Caffe2 8-bit dequantization operators (e.g., SparseLengthSum\footnote{\url{https://caffe2.ai/docs/operators-catalogue.html\#sparselengthssum}}) that are already heavily optimized.     
\end{itemize} 

We empirically demonstrate the effectiveness of our proposed quantization approaches on DNN-based recommendation models~\cite{zhang2018training, naumov2019deep} and also a production model in Facebook. The results show that the proposed approaches consistently outperform state-of-the-art post-training quantization approaches in reducing quantization error and accuracy degradation. Row-wise uniform quantization with greedy search can reduce the model size to 13.3\%-25.0\% of the baseline single-precision models with negligible accuracy loss. Codebook-based quantization can reduce the model size to 18.5\%-37.5\% of the single-precision model with no accuracy loss. We deploy our uniform quantization technique on a production model in Facebook and demonstrate that it can reduce the model size to only 13.89\% of the single-precision version while the model quality stays neutral. 



\section{Prior Quantization Methods and Their Limitations}
Let $x$ be a value clipped to the range $[x_{min}, x_{max}]$. Quantization using $n$ bits maps $x$ to an integer in the range $[0, 2^n-1]$, where each integer corresponds to a quantized value. If the quantized values are a set of discrete, evenly-spaced grid points, the methods are called \textit{uniform quantization}. Otherwise, they are called \textit{non-uniform quantization}. This section reviews several state-of-the-art post-training uniform quantization methods and explains their limitations on embedding table quantization. 


Let $x_{int}$ and $x_{float}$ be the quantized and dequantized values respectively. 
Uniform quantization proceeds as follows:
\begin{align}
    x_{int} = round\left(\frac{x - x_{min}}{x_{max} - x_{min}} * (2^n - 1)\right) = round\left(  \frac{x - bias}{scale} \right),
    \label{eq:quant}
\end{align}
where $ scale = \frac{x_{max} - x_{min}}{2^n - 1}$ and $bias =  x_{min}$.
The de-quantization operation is: $x_{float} = scale * x_{int} + bias$. The quantization is \textit{symmetric} if $x_{max} = - x_{min}$. Otherwise, it is  \textit{asymmetric}. To ease the description below, we define a quantization function\footnote{An alternative uniform quantization uses: $x_{int}=round(x/scale) - zero\_point$. This method is better when the inputs to quantize have lots of zeros, e.g., ReLU activations. We found that the mapping in Eq.~\ref{eq:quant} provides better accuracy for embedding table quantization. } as: $x_{float} = Q(x, x_{min}, x_{max})$. 

\begin{wrapfigure}{r}{0.5\textwidth}
  \begin{center}
    \includegraphics[width=0.4999\textwidth]{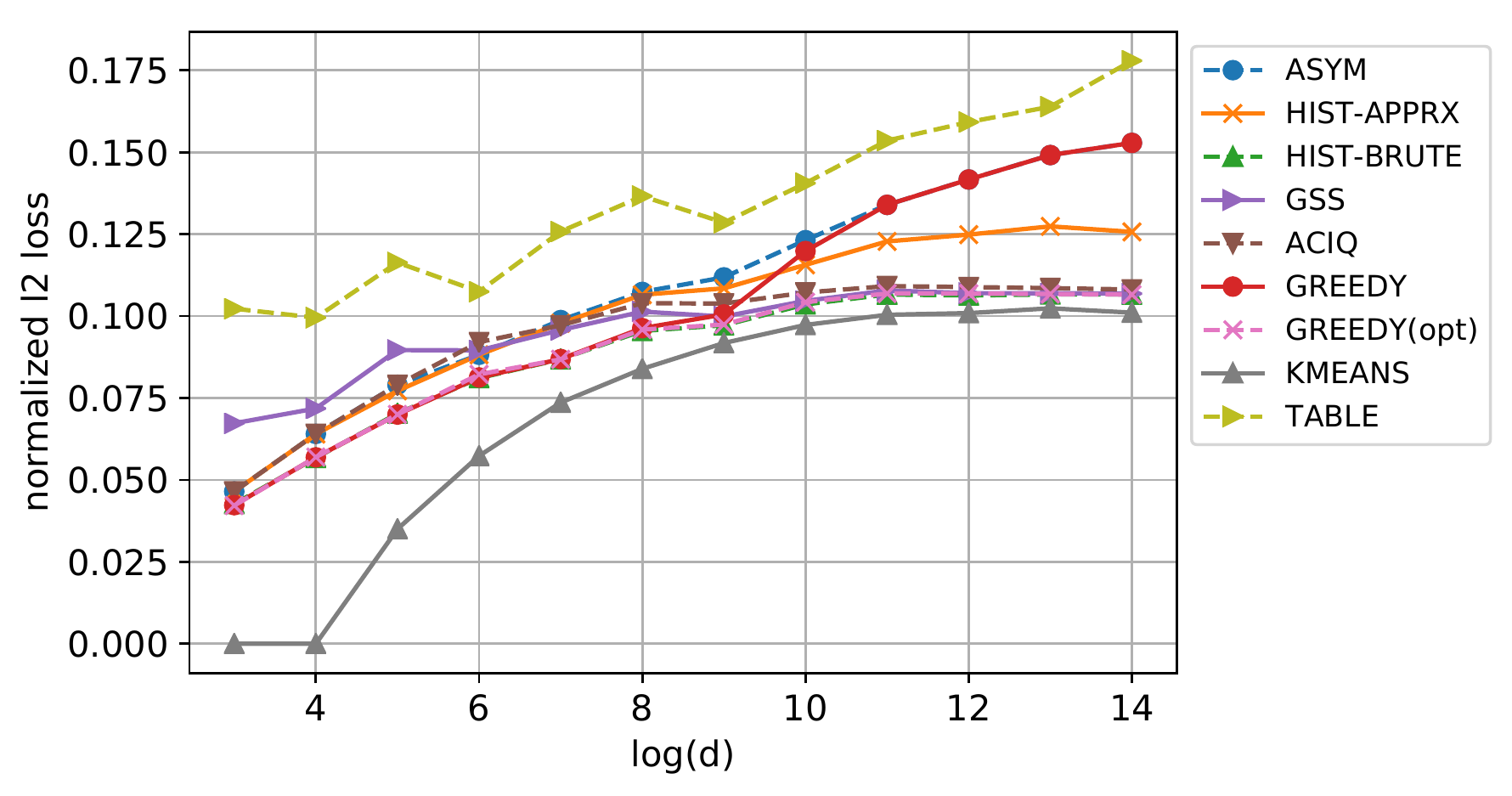}
  \end{center}
  \caption{\small{The normalized $\ell_2$ loss of 4-bit quantization with different embedding dimensions on a FP32 embedding table with 10 row vectors. The values in the embedding table are randomly sampled from a normal distribution, which is in favor of GSS and especially ACIQ. TABLE applies range-based uniform quantization on the entire table while the other methods are on row vectors. HIST-APPRX and HIST-BRUTE use $b=200$, GREEDY uses $b=200, r=0.16$; GREEDY (opt) uses $b=1000, r=0.5$.}}
  \label{fig:mse_diff_d}
\end{wrapfigure}

Without loss of generality, let $X\in \mathbb{R}^N$ be an input vector to quantize. Tensors with higher dimensions can be flattened to a vector. Because each value is scaled by $x_{max} - x_{min}$, the naive quantization using $x_{max} = max(X)$ and $x_{min} = min(X)$ is sensitive to outliers, i.e., values with large magnitude in the input $X$, and could cause large accuracy drops. We refer to this method \textbf{range-based asymmetric quantization (ASYM)}.


State-of-the-art post-training quantization techniques rely on the clipping to mitigate accuracy degradation.  They differ in their way to minimize the mean squared error (MSE) of the original values and the quantized values: 
\begin{equation}
    f({x_{min}, x_{max}}) = \|X - Q(X, x_{min}, x_{max})\|_2^2.  \label{eq:mse_minimization}
\end{equation}

\paragraph{Histogram-based Quantization (HIST)}
This method chooses the clipping thresholds which minimize the MSE between the histogram of floating-point inputs and that of the quantized versions~\cite{shin2016fixed}. Let $x_i$ and $h(x_i)$, where $i=1, \cdots, b$, be the bin value and the frequency of the $i$-th bin in the inputs' histogram. The optimization objective is defined as: 
$f_{hist}(x_{min}, x_{max}) = \frac{1}{b} \sum_{i=1}^{b} h(x_i) * (x_i - Q(x_i, x_{min}, x_{max}))^2$.
We used the approximate algorithm (\textbf{HIST-APPRX}) implemented in Caffe2~\cite{caffe2_norm_minimization} that scales linearly with the number of bins to solve the above optimization problem. We also implemented a brute force approach (\textbf{HIST-BRUTE}) to find better solutions. Its time complexity is $O(b^3)$ (See Appendix~A). 

\paragraph{ACIQ}
Analytical Clipping for Integer Quantization (ACIQ)~\cite{banner2018aciq} derives the clipping thresholds analytically from the distribution of the tensor. It assumes that the values in the tensor are sampled from a Gaussian or a Laplacian distribution. After determining the distribution to use, it uses an approximate closed-form solution for the clip thresholds which minimizes MSE in Eq.~\ref{eq:mse_minimization}. For example, if the tensor is closer to a Laplacian distribution, the clipping thresholds for 4-bit quantization are calculated using the formula:
$ x_{min} = \mathbb{E}(X) - \alpha, x_{max} = \mathbb{E}(X) + \alpha$, where $\alpha = 5.03 * \mathbb{E}(|X - \mathbb{E}(X))$. We used the open-source code from the authors\footnote{\url{https://github.com/submission2019/cnn-quantization}}. 

\paragraph{Quantization with Golden Section Search (GSS)}
Instead of approximating the floating-point inputs using a histogram or assuming it follows a certain distribution, this approach finds a range limit $x_{thr}$ that minimizes MSE using golden section search (GSS)~\cite{kiefer1953sequential} for symmetric quantization. The objective function is simplified as:  $f_{sym}(x_{thr}) = \frac{1}{N} \|X - Q(X, - x_{thr}, x_{thr}))\|_2^2$. The method is applied to compress word embeddings in~\cite{jian2019embeddings}.

\paragraph{Their Limitations} Quantization on embedding tables is commonly applied to row vectors to reduce the quantization error (See  ASYM v.s. TABLE in Figure~\ref{fig:mse_diff_d}). The embedding dimension in recommendation models is usually 8 to 200~\cite{naumov2019deep}.  The above clipping-based approaches are better than the range-based asymmetric quantization (ASYM) method when quantizing weights and activations of CNNs to 4-bit. However, we empirically observed that they are no better than ASYM when the input $X$ to quantize has a small dimension (i.e., a small number of values), as in the case of embedding table quantization. Approximating row vectors in an embedding table using histograms or distributions could give large quantization error.

Figure~\ref{fig:mse_diff_d} shows the normalized $\ell_2$ loss of different quantization methods with various embedding dimensions. Normalized $\ell_2$ loss is calculated as ${\|X - Q(X, x_{min}, x_{max})\|_2}/{\|X\|_2}$. It measures relative quantization errors. Overall, when the embedding dimension is larger than 1024, GSS, ACIQ, HIST-APPRX, and HIST-BRUTE can achieve a smaller loss compared with ASYM. However, when the embedding dimension is small (e.g. 64), the advantage of GSS, ACIQ, and HIST-APPRX over ASYM is gone. GSS is even much worse than ASYM. Although HIST-BRUTE is still better than ASYM, it is very time-consuming (millions of times slower than ASYM) and too expensive to be applied in real-world recommendation applications that require continuous learning and thus periodic quantization for model serving (See Appendix~A). 


\section{Proposed Quantization Approaches}
This section elaborates the proposed uniform quantization with greedy search and codebook-based quantization with k-means clustering.

\paragraph{Uniform Quantization with Greedy Search (GREEDY)}

\begin{wrapfigure}{R}{0.5\textwidth}
\begin{minipage}{0.5\textwidth}
\begin{algorithm}[H]
\small 
\caption{greedy search}
\label{alg:greedy}
\begin{algorithmic}[1]
\REQUIRE X $\quad$\COMMENT a vector to quantize.
\REQUIRE b $\quad$ \COMMENT default: 200
\REQUIRE r $\quad$ \COMMENT default: 0.16 
\ENSURE xmin, xmax \COMMENT range used for quantization
\STATE xmin = cur\_min = min(X)
\STATE xmax = cur\_max = max(X)
\STATE loss = compute\_loss(X, Q(X, xmin, xmax))
\STATE stepsize = (xmax - xmin)/b
\STATE min\_steps = b * (1 - r) * stepsize  
\WHILE{ cur\_min + min\_steps < cur\_max }
\STATE loss\_l = compute\_loss(X, Q(X, cur\_min + stepsize, cur\_max))
\STATE   loss\_r = compute\_loss(X,Q(X, cur\_min, cur\_max - stepsize))
\IF   {loss\_l < loss\_r}
\STATE        cur\_min = cur\_min + stepsize
\IF {loss\_l < loss}
\STATE loss, xmin = loss\_1, cur\_min 
\ENDIF 
\ELSE
\STATE cur\_max = cur\_max - stepsize
\IF {loss\_r < loss}
\STATE loss, xmax = loss\_r, cur\_max
\ENDIF 

\ENDIF 
\ENDWHILE 
\RETURN xmin, xmax 

\end{algorithmic}
\end{algorithm}
\end{minipage}
\end{wrapfigure}

 To overcome the limitations of the prior uniform quantization approaches, we propose a greedy search algorithm (see Algorithm~\ref{alg:greedy}) to find the optimal clipping thresholds. 
 The algorithm is inspired by the 1-D golden section search (GSS) and directly targets at minimizing the MSE objective function in Eq.~\ref{eq:mse_minimization}. 
Although 2-D GSS was proposed recently, it is not applicable in general as it is too consuming~\cite{chang2009n}. 
 The basic idea of greedy search is to find as many local optima as possible and select the best one as the clipping thresholds. The algorithm takes the input vector $X$ and two hyperparameters $b$ and $r$ that balance the optimality of its solution and its time complexity.

The algorithm initializes $xmin$ and $xmax$ with the range of the input $X$ (lines 1-2). It then gradually increases $xmin$ or decreases $xmax$  by one $stepsize$ to reduce the range and find a smaller $loss$ calculated as Eq.~\ref{eq:mse_minimization} (lines 7-16). The algorithm stops when the current range is $1-r$ percentage of the range of $X$ (lines 5-6). The larger the $b$ and $r$ are, the better the found solution will be but the higher the time cost is ($O(b\times r)$ time complexity). The default value of $b$ and $r$ are set as $200$ and $0.16$ respectively.


\paragraph{Codebook-based Quantization}
Codebook-based quantization is a type of non-uniform quantization that maps each input value to the index of a value in the codebook. Quantizing a value to 4 bits means the number of values in a codebook cannot be larger than 16. We consider the following two codebook-based quantization variants: \textbf{ Quantization with Rowwise Clustering (KMEANS)} and \textbf{Quantization with Two-Tier Clustering (KMEANS-CLS)}. KMEANS algorithm applies k-means clustering to produce a 16-value codebook for each row vector, and then maps each value in the row vector to the index of the codebook based on its cluster assignment. Although the algorithm has less model size reduction than uniform quantization due to the storage overhead of codebooks, it has the potential to achieve a lower MSE and avoid accuracy degradation.

 To achieve a larger compression rate, KMEANS-CLS applies k-means clustering in a more coarse-grained way. The algorithm first groups similar row vectors in an embedding table into blocks (called tier-1 clustering) and then generates a 16-value codebook for each block (called tier-2 clustering). Both steps use k-means clustering. Let $K$ be the number of clusters in tier-1 clustering. After 4-bit quantization using KMEANS-CLS, the required number of bytes to store a $N\times d$ embedding table is $Nd/2 + N\log_2K/8 + 64K$, where $\log_2K/8$ is the number of bytes used to store tier-1 cluster assignment.

\section{Efficient 4-bit Embedding Operation Implementation}

\begin{table}[]
\caption{Computational throughput in billion sums per second for SparseLengthsSum operators.
The performance is measured using a single core of Intel Xeon Gold 6138 CPU @ 2.0 GHz with turbo-mode off.}
\label{tab:speed}
\centering 
\scriptsize 
\tabcolsep=0.1cm 
\begin{tabular}{|l||l|l|l|l||l|l|l|l|}
\hline
\multirow{2}{*}{Data type} & \multicolumn{4}{l|}{Cache non-resident case} & \multicolumn{4}{l|}{Cache resident case} \\ \cline{2-9} 
 & d=64 & d=128 & d=256 & d=512 & d=64 & d=128 & d=256 & d=512 \\ \hline\hline
FP32 & 1.939 & 1.908 & 1.997 & 2.063 & 2.804 & 4.165 & 4.209 & 4.127 \\ \hline
INT8 & 1.246 & 1.511 & 2.726 & 3.076 & 2.242 & 2.510 & 3.748 & 3.450 \\ \hline
INT4 & 1.608 & 2.047 & 2.532 & 5.581 & 2.093 & 2.878 & 6.454 & 6.893 \\ \hline
\end{tabular}
\end{table}

A challenge for quantizing embedding tables in less than 8-bit is the overhead of dequantization when reading the tables.
This is because, in most commonly used processors, 8-bit is the smallest granularity in which instructions can operate with, and less than 8-bit granularity requires bit manipulations like or, shift, and so on.
Nevertheless, we found that we can sustain good enough dequantization throughput with careful use of vector instructions available in recent CPUs (e.g., AVX512 in Intel Skylake CPUs) as shown in Table~\ref{tab:speed}.
We measure computational throughput of SparseLengthsSum operator (the most time-consuming operator reading embedding tables in our recommendation models~\cite{naumov2019deep}) in FP32, INT8, and INT4, both in cache non-resident and cache resident cases.
In cache non-resident cases, we flush the last level cache between benchmark runs, which is more representative of running big recommendation models with many huge embedding tables.
The cache resident cases are to see upper bound computational throughputs (a worst case for 4-bit embedding tables).
We can see that in most cases the speed of 4-bit SparseLengthsSum is on par or faster than its highly-optimized 8-bit or FP32 counterparts running in production.

\section{Experiments}
\label{sec:eval} 

In this section, we present the experimental results of the proposed approaches. We use the Terabyte Criteo data~\cite{dataset}. It is a click prediction dataset that has a size of 1.3TB and contains more than 4.3 billion records. The dataset is a commonly used benchmark dataset for ranking applications.

The ranking problem is a binary classification problem. The models we used are DNN models~\cite{naumov2019deep}. For categorical features, following the same procedure as in \cite{zhang2018training}, we transform them into dense vectors using embedding tables. The number of rows in embedding tables corresponds to the number of categorical features with a maximum of 50 million. The number of columns corresponds to the embedding dimension. We choose a variety of embedding dimensions: 8, 16, 32, 64, and 128, that are commonly used in ranking models. The dense embeddings of the categorical features, concatenated with the dense vector formed by dense features, are taken as the input to a neural network with 2 fully-connected (FC) layers. The FC layers have a width of 512.   
The models are trained using Adagrad~\cite{duchi2011adaptive} with a batch size of 100. The initial learning rate is set to 0.015 for embedding tables and 0.005 for the rest of the parameters. All the parameters are trained using single-precision (FP32). All the embedding tables are quantized to use 4 bits after model training is finished. 

We compare the proposed approaches (GREEDY, KMEANS, KMEANS-CLS) with other uniform quantization approaches including SYM, GSS, ASYM, HIST-APPROX, HIST-BRUTE, ACIQ. Because k-means is sensitive to initialization, we initialize cluster centers 
using uniform quantization results from ASYM.
The default hyperparameter settings ($b=200, r=0.16$) are used for greedy search. HIST-APPRX uses $b=200$ as it gives the best performance after a grid-based hyper-parameter tuning. For KMEANS-CLS,
we choose the $K$ such that it achieves the same compression rate as the uniform quantization approaches. If a method is appended with ``(FP16)'', it means that the scales and biases in uniform quantization and the codebooks in codebook-based quantization are stored using FP16 instead of FP32.
Besides the baseline where embedding tables are not quantized (FP32), we include another baseline that uses range-based uniform quantization to quantize all embedding tables to 8 bits (ASYM-8BITS). 
%
We evaluate the performance of the quantization approaches using three evaluation metrics: Normalized $\ell_2$ loss, model log loss, and model size.

\begin{table}[t]
\caption{Normalized $l2$ loss with different quantization methods and embedding dimensions. }
\label{tab:l2_loss}
\scriptsize  
\centering 
\tabcolsep=0.1cm 
\begin{tabular}{|l|l||l|l|l|l|l|}
\hline
Methods & Description &  d=8 & d=16 & d=32 & d=64 & d=128 \\ \hline\hline
ASYM-8BITS & Asymmetric, $x_{min}=min(X), x_{max}=max(X)$ & 0.00260 & 0.00329 & 0.00376 & 0.00387 & 0.00400 \\ \hline\hline
SYM & Symmetric, $x_{min} = -x_{max}, x_{max} = max(|X|)$ & 0.05564 & 0.06296 & 0.06785 & 0.06836 & 0.06928 \\ \hline
GSS & Symmetric with Golden Section Search & 0.05269 & 0.05965 & 0.06328 & 0.06400 & 0.06423 \\ \hline
ASYM & Asymmetric, $x_{min}=min(X), x_{max}=max(X)$ & 0.04451 & 0.05479 & 0.06387 & 0.06608 & 0.06781 \\ \hline
HIST-APPRX & Asymmetric with histogram-based approximation~\cite{caffe2_norm_minimization} & 0.04452 & 0.05512 & 0.06409 & 0.06589 & 0.06768 \\ \hline
HIST-BRUTE & Asymmetric with histogram-based brute force algorithm & 0.04156 & 0.05082 & 0.05881 & 0.06083 & 0.06272 \\ \hline
ACIQ & Analytical Clipping for Integer Quantization~\cite{banner2018aciq}& 0.04451 & 0.05479 & 0.06387 & 0.06742 & 0.07665 \\ \hline
\textbf{GREEDY} & Asymmetric with greedy search (ours) & \textbf{0.03889} & \textbf{0.04878} & \textbf{0.05744} & \textbf{0.05991} & \textbf{0.06221} \\ \hline
\textbf{GREEDY (FP16)} & Asymmetric with greedy search (ours) & \textbf{0.03889} & \textbf{0.04879} & \textbf{0.05744} & \textbf{0.05991} & \textbf{0.06221} \\ \hline\hline
KMEANS-CLS (FP16) & Two-tier k-means clustering with uniform init (ours) &  0.03948 & 0.05349 & 0.06826 & 0.07369 & 0.07287 \\ \hline
\textbf{KMEANS (FP16)} & Rowwise kmeans clustering with uniform init (ours) & \textbf{0} & \textbf{0} & \textbf{0.03670} & \textbf{0.05160} & \textbf{0.05781} \\ \hline
\end{tabular} \\
\vspace{-0.1cm} 
\begin{flushleft} "-8BITS": 8-bit quantization; otherwise, 4-bit quantization. "(FP16)": scales and biases or values of a codebook in FP16; otherwise, FP32.  \end{flushleft} 
\vspace{-0.4cm} 
\end{table}

Table~\ref{tab:l2_loss} lists the normalized $\ell_2$ loss results on an embedding table from models with different embedding dimensions. Overall, our proposed approach GREEDY consistently gives the smallest loss among all 4-bit uniform quantization approaches. Using FP16 for scales and biases further reduces the embedding table size without affecting the loss. KMEANS achieves the smallest normalized $\ell_2$ loss for the use of codebook. Even though KMEANS-CLS variants can achieve the same compression rate as the uniform quantization approaches, they suffer from larger losses, indicating the importance of row-wise quantization for embedding tables.

\begin{table}[t]
\scriptsize
\tabcolsep=0.1cm
\centering 
\caption{Model log loss and size with different quantization methods and embedding dimensions. 
}
\label{tab:model_loss}
\begin{tabular}{|l||l|l|l|l|l|l|l|l|l|l|}
\hline
\multirow{2}{*}{Methods} & \multicolumn{2}{l|}{d=8} & \multicolumn{2}{l|}{d=16} & \multicolumn{2}{l|}{d=32} & \multicolumn{2}{l|}{d=64} & \multicolumn{2}{l|}{d=128} \\ \cline{2-11} 
 & loss & size & loss & size & loss & size & loss & size & loss & size \\ \hline\hline
FP32 (no quantization) & 0.12522 & 8.07GB & 0.12489 & 16.14GB & 0.12468 & 32.27GB & 0.12451 & 64.54GB & 0.12454 & 129.09GB \\ \hline\hline
ASYM-8BITS & 0.12522 & 49.98\% & 0.12489 & 37.49\% & 0.12469 & 31.25\% & 0.12451 & 28.12\% & 0.12454 & 26.56\% \\ \hline\hline
SYM & 0.12528 & \multirow{7}{*}{37.49\%} & 0.12507 & \multirow{7}{*}{24.99\%} & 0.13266 & \multirow{7}{*}{18.75\%} & 0.13107 & \multirow{7}{*}{15.62\%} & 0.12470 & \multirow{7}{*}{14.06\%} \\ \cline{1-2} \cline{4-4} \cline{6-6} \cline{8-8} \cline{10-10}
GSS & 0.12527 &  & 0.12504 &  & 0.13199 &  & 0.12843 &  & 0.12459 &  \\ \cline{1-2} \cline{4-4} \cline{6-6} \cline{8-8} \cline{10-10}
ASYM & 0.12526 &  & 0.12491 &  & 0.12496 &  & 0.12494 &  & 0.12455 &  \\ \cline{1-2} \cline{4-4} \cline{6-6} \cline{8-8} \cline{10-10}
HIST-APPRX & 0.12525 &  & 0.12492 &  & 0.12497 &  & 0.12498 &  & 0.12455 &  \\ \cline{1-2} \cline{4-4} \cline{6-6} \cline{8-8} \cline{10-10}
HIST-BRUTE & 0.12525 &  & 0.12490 &  & 0.12490 &  & 0.12489 &  & 0.12454 &  \\ \cline{1-2} \cline{4-4} \cline{6-6} \cline{8-8} \cline{10-10}
ACIQ & 0.12526 &  & 0.12491 &  & 0.12804 &  & 0.12514 &  & 0.12455 &  \\ \cline{1-2} \cline{4-4} \cline{6-6} \cline{8-8} \cline{10-10}
GREEDY & \textbf{0.12525} &  & \textbf{0.12490} &  & \textbf{0.12489} &  & \textbf{0.12485} &  & \textbf{0.12454} &  \\ \hline
\textbf{GREEDY (FP16)} & \textbf{0.12525} & \textbf{24.99\%} & \textbf{0.12490} & \textbf{18.74\%} & \textbf{0.12489} & \textbf{15.62\%} & \textbf{0.12485} & \textbf{14.06\%} & \textbf{0.12454} & \textbf{13.28\%} \\ \hline
\textbf{KMEANS (FP16)} & \textbf{-} & \textbf{-} & \textbf{-} & \textbf{-} & \textbf{0.12469} & \textbf{37.50\%} & \textbf{0.12451} & \textbf{25.00\%} & \textbf{0.12454} & \textbf{18.75\%} \\ \hline
\end{tabular}
\vspace{-0.2cm} 
\end{table}

Table~\ref{tab:model_loss} lists the model log loss and model size for the models after 4-bit quantization. Overall, GREEDY consistently gives the smallest model log loss compared with other uniform quantization approaches while reducing the models to 13.3\%-25.0\% of the single-precision model size. The proposed KMEANS approach can even get the same model loss as the original single-precision model while reducing the models to 18.8\%-37.5\% of the single-precision model size.  

We deployed GREEDY on one of the ranking applications at Facebook. The application uses a DNN model trained on billions of records. Being able to reduce the model size using post-training 4-bit quantization while preserving model accuracy is a challenging task. 
Our experimental results show that the 4-bit uniform quantization with greedy search can reduce the model size to only 13.89\% of the single-precision version while the model quality stays neutral. This demonstrates the practicality of our approach in real applications. 

\section{Conclusions and Future Work}
We proposed row-wise uniform quantization with greedy search and non-uniform quantization with k-means clustering to improve 4-bit post-training quantization on embedding tables.
We empirically showed that the proposed approaches consistently outperform state-of-the-art quantization methods on reducing the quantization error and the model accuracy degradation. 
The model size reduction resulting from 4-bit quantization makes it possible to use even larger embedding tables for potentially better model accuracy. In the future, we want to explore how much accuracy gain can be achieved by increasing model size while applying 4-bit quantization to meet a certain space budget.

\bibliographystyle{plain}
\bibliography{paper}

\begin{thebibliography}{10}

\bibitem{caffe2_norm_minimization}
caffe2 histogram-based norm minimization.
\newblock
  \url{https://caffe2.ai/doxygen-c/html/norm__minimization_8cc_source.html}.
\newblock Accessed: 2019-08-03.

\bibitem{dataset}
Criteo releases industry’s largest-ever dataset for machine learning to
  academic community.
\newblock
  \url{https://www.criteo.com/news/press-releases/2015/07/criteo-releases-industrys-largest-ever-dataset/}.
\newblock Accessed: 2019-08-03.

\bibitem{banner2018aciq}
Ron Banner, Yury Nahshan, Elad Hoffer, and Daniel Soudry.
\newblock Aciq: Analytical clipping for integer quantization of neural
  networks.
\newblock {\em arXiv preprint arXiv:1810.05723}, 2018.

\bibitem{chang2009n}
Yen-Ching Chang.
\newblock N-dimension golden section search: Its variants and limitations.
\newblock In {\em 2009 2nd International Conference on Biomedical Engineering
  and Informatics}, pages 1--6. IEEE, 2009.

\bibitem{cheng2016wide}
Heng-Tze Cheng, Levent Koc, Jeremiah Harmsen, Tal Shaked, Tushar Chandra,
  Hrishi Aradhye, Glen Anderson, Greg Corrado, Wei Chai, Mustafa Ispir, et~al.
\newblock Wide \& deep learning for recommender systems.
\newblock In {\em Proceedings of the 1st workshop on deep learning for
  recommender systems}, pages 7--10. ACM, 2016.

\bibitem{choukroun2019low}
Yoni Choukroun, Eli Kravchik, and Pavel Kisilev.
\newblock Low-bit quantization of neural networks for efficient inference.
\newblock {\em arXiv preprint arXiv:1902.06822}, 2019.

\bibitem{courbariaux1602binarynet}
Matthieu Courbariaux and Yoshua Bengio.
\newblock Binarynet: Training deep neural networks with weights and activations
  constrained to+ 1 or- 1. arxiv 2016.
\newblock {\em arXiv preprint arXiv:1602.02830}.

\bibitem{de2018high}
Christopher De~Sa, Megan Leszczynski, Jian Zhang, Alana Marzoev, Christopher~R
  Aberger, Kunle Olukotun, and Christopher R{\'e}.
\newblock High-accuracy low-precision training.
\newblock {\em arXiv preprint arXiv:1803.03383}, 2018.

\bibitem{duchi2011adaptive}
John Duchi, Elad Hazan, and Yoram Singer.
\newblock Adaptive subgradient methods for online learning and stochastic
  optimization.
\newblock {\em Journal of Machine Learning Research}, 12(Jul):2121--2159, 2011.

\bibitem{goncharenko2018fast}
Alexander Goncharenko, Andrey Denisov, Sergey Alyamkin, and Evgeny Terentev.
\newblock Fast adjustable threshold for uniform neural network quantization.
\newblock {\em arXiv preprint arXiv:1812.07872}, 2018.

\bibitem{he2016effective}
Qinyao He, He~Wen, Shuchang Zhou, Yuxin Wu, Cong Yao, Xinyu Zhou, and Yuheng
  Zou.
\newblock Effective quantization methods for recurrent neural networks.
\newblock {\em arXiv preprint arXiv:1611.10176}, 2016.

\bibitem{he2017neural}
Xiangnan He, Lizi Liao, Hanwang Zhang, Liqiang Nie, Xia Hu, and Tat-Seng Chua.
\newblock Neural collaborative filtering.
\newblock In {\em Proceedings of the 26th international conference on world
  wide web}, pages 173--182. International World Wide Web Conferences Steering
  Committee, 2017.

\bibitem{kiefer1953sequential}
Jack Kiefer.
\newblock Sequential minimax search for a maximum.
\newblock {\em Proceedings of the American mathematical society},
  4(3):502--506, 1953.

\bibitem{krishnamoorthi2018quantizing}
Raghuraman Krishnamoorthi.
\newblock Quantizing deep convolutional networks for efficient inference: A
  whitepaper.
\newblock {\em arXiv preprint arXiv:1806.08342}, 2018.

\bibitem{li2017training}
Hao Li, Soham De, Zheng Xu, Christoph Studer, Hanan Samet, and Tom Goldstein.
\newblock Training quantized nets: A deeper understanding.
\newblock In {\em Advances in Neural Information Processing Systems}, pages
  5811--5821, 2017.

\bibitem{lian2018xdeepfm}
Jianxun Lian, Xiaohuan Zhou, Fuzheng Zhang, Zhongxia Chen, Xing Xie, and
  Guangzhong Sun.
\newblock xdeepfm: Combining explicit and implicit feature interactions for
  recommender systems.
\newblock In {\em Proceedings of the 24th ACM SIGKDD International Conference
  on Knowledge Discovery \& Data Mining}, pages 1754--1763. ACM, 2018.

\bibitem{jian2019embeddings}
Avner May, Jian Zhang, Tri Dao, and Christopher Ré.
\newblock On the downstream performance of compressed word embeddings.
\newblock {\em arXiv preprint arXiv:1909.01264}, 2019.

\bibitem{micikevicius2017mixed}
Paulius Micikevicius, Sharan Narang, Jonah Alben, Gregory Diamos, Erich Elsen,
  David Garcia, Boris Ginsburg, Michael Houston, Oleksii Kuchaiev, Ganesh
  Venkatesh, et~al.
\newblock Mixed precision training.
\newblock {\em arXiv preprint arXiv:1710.03740}, 2017.

\bibitem{migacz20178}
Szymon Migacz.
\newblock 8-bit inference with tensorrt.
\newblock In {\em GPU technology conference}, volume~2, page~7, 2017.

\bibitem{mikolov2013distributed}
Tomas Mikolov, Ilya Sutskever, Kai Chen, Greg~S Corrado, and Jeff Dean.
\newblock Distributed representations of words and phrases and their
  compositionality.
\newblock In {\em Advances in neural information processing systems}, pages
  3111--3119, 2013.

\bibitem{naumov2019deep}
Maxim Naumov, Dheevatsa Mudigere, Hao-Jun~Michael Shi, Jianyu Huang, Narayanan
  Sundaraman, Jongsoo Park, Xiaodong Wang, Udit Gupta, Carole-Jean Wu,
  Alisson~G Azzolini, et~al.
\newblock Deep learning recommendation model for personalization and
  recommendation systems.
\newblock {\em arXiv preprint arXiv:1906.00091}, 2019.

\bibitem{park2018deep}
Jongsoo Park, Maxim Naumov, Protonu Basu, Summer Deng, Aravind Kalaiah, Daya
  Khudia, James Law, Parth Malani, Andrey Malevich, Satish Nadathur, et~al.
\newblock Deep learning inference in facebook data centers: Characterization,
  performance optimizations and hardware implications.
\newblock {\em arXiv preprint arXiv:1811.09886}, 2018.

\bibitem{pennington2014glove}
Jeffrey Pennington, Richard Socher, and Christopher Manning.
\newblock Glove: Global vectors for word representation.
\newblock In {\em Proceedings of the 2014 conference on empirical methods in
  natural language processing (EMNLP)}, pages 1532--1543, 2014.

\bibitem{shin2016fixed}
Sungho Shin, Kyuyeon Hwang, and Wonyong Sung.
\newblock Fixed-point performance analysis of recurrent neural networks.
\newblock In {\em 2016 IEEE International Conference on Acoustics, Speech and
  Signal Processing (ICASSP)}, pages 976--980. IEEE, 2016.

\bibitem{sung2015resiliency}
Wonyong Sung, Sungho Shin, and Kyuyeon Hwang.
\newblock Resiliency of deep neural networks under quantization.
\newblock {\em arXiv preprint arXiv:1511.06488}, 2015.

\bibitem{zhang2018training}
Jian Zhang, Jiyan Yang, and Hector Yuen.
\newblock Training with low-precision embedding tables.
\newblock In {\em Systems for Machine Learning Workshop at NeurIPS}, volume
  2018, 2018.

\bibitem{zhao2019improving}
Ritchie Zhao, Yuwei Hu, Jordan Dotzel, Chris De~Sa, and Zhiru Zhang.
\newblock Improving neural network quantization without retraining using
  outlier channel splitting.
\newblock In {\em International Conference on Machine Learning}, pages
  7543--7552, 2019.

\bibitem{zhou2016dorefa}
Shuchang Zhou, Yuxin Wu, Zekun Ni, Xinyu Zhou, He~Wen, and Yuheng Zou.
\newblock Dorefa-net: Training low bitwidth convolutional neural networks with
  low bitwidth gradients.
\newblock {\em arXiv preprint arXiv:1606.06160}, 2016.

\end{thebibliography}

\newpage 
\appendix

\section{HIST-BRUTE}
\label{appendix:time_complexity}
In this section, we present the algorithm and the complexity analysis of the brute force histogram-based quantization approach (HIST-BRUTE). Its pseudo-code is shown in Algorithm~\ref{alg:hist-brute}. For 4-bit quantization, the algorithm uses a histogram with 16 number of bins $dst\_nbins$ to approximate the histogram of the inputs with $b$ number of bins. Lines 1-9 are for initialization. The algorithm selects different numbers of consecutive bins from the inputs' histogram and approximates these selected bins using 16 target bins (lines 10-36). The selected bins determine the clipping thresholds (lines 37-42). HIST-BRUTE has a time complexity of $O(b^3)$. 

\begin{algorithm}[H]
\small 
\caption{HIST-BRUTE}
\label{alg:hist-brute}
\begin{algorithmic}[1]
\REQUIRE X \COMMENT a vector to quantize.
\REQUIRE b \COMMENT number of bins used to generate histogram, default: 200
\ENSURE xmin, xmax \COMMENT range used for quantization
\STATE \COMMENT{Initialize}
\STATE xmin = min(X)
\STATE xmax = max(X)
\STATE histogram = get\_histogram(X, b)
\STATE dst\_nbins = 16 
\STATE bin\_width = (xmax - xmin)/b
\STATE norm\_min = $\infty$
\STATE best\_start\_bin = -1
\STATE best\_nbins\_selected = 1
\FOR{nbins\_selected = 1 \TO b}
\STATE start\_bin\_begin = 0
\STATE start\_bin\_end = b - nbins\_selected + 1
\STATE dst\_bin\_width = bin\_width * nbins\_selected / (dst\_nbins - 1)
\FOR{start\_bin = start\_bin\_begin \TO start\_bin\_end}
\STATE norm = 0
\STATE \COMMENT{Go over each histogram bin and accumulate errors.}
\FOR{src\_bin = 0 \TO b} 
\STATE src\_bin\_begin = (src\_bin - start\_bin) * bin\_width
\STATE src\_bin\_end = src\_bin\_begin + bin\_width
\STATE \COMMENT{Determine which dst\_bins the beginning and end of src\_bin belong to}
\STATE dst\_bin\_of\_begin = min(dst\_nbins - 1, max(0, floor((src\_bin\_begin + 0.5 * dst\_bin\_width) / dst\_bin\_width)))
\STATE dst\_bin\_of\_end = min(dst\_nbins - 1, max(0, floor((src\_bin\_end + 0.5 * dst\_bin\_width) / dst\_bin\_width)))
\STATE dst\_bin\_of\_begin\_center = dst\_bin\_of\_begin * dst\_bin\_width
\STATE density = histogram[src\_bin] / bin\_width
\STATE delta\_begin = src\_bin\_begin - dst\_bin\_of\_begin\_center
\IF{dst\_bin\_of\_begin == dst\_bin\_of\_end}
\STATE delta\_end = src\_bin\_end - dst\_bin\_of\_begin\_center
\STATE norm += get\_l2\_norm(delta\_begin, delta\_end, density)
\ELSE  \STATE delta\_end = dst\_bin\_width / 2
\STATE norm += get\_l2\_norm(delta\_begin, delta\_end, density)
\STATE norm += (dst\_bin\_of\_end - dst\_bin\_of\_begin - 1) * get\_l2\_norm( -dst\_bin\_width / 2, dst\_bin\_width / 2, density)
\STATE dst\_bin\_of\_end\_center = dst\_bin\_of\_end * dst\_bin\_width
\STATE delta\_begin = -dst\_bin\_width / 2
\STATE delta\_end = src\_bin\_end - dst\_bin\_of\_end\_center
\STATE norm += get\_l2\_norm(delta\_begin, delta\_end, density)
\ENDIF 
\ENDFOR 
\IF{ norm < norm\_min}
\STATE norm\_min = norm
\STATE best\_start\_bin = start\_bin
\STATE best\_nbins\_selected = nbins\_selected
\ENDIF 
\ENDFOR 
\ENDFOR 

\STATE xmin = xmin + bin\_width * best\_start\_bin
\STATE xmax = xmax + bin\_width * (best\_start\_bin + best\_nbins\_selected)
\RETURN xmin, xmax 

\end{algorithmic}
\end{algorithm}

\begin{figure}
  \begin{center}
    \includegraphics[width=0.6\textwidth]{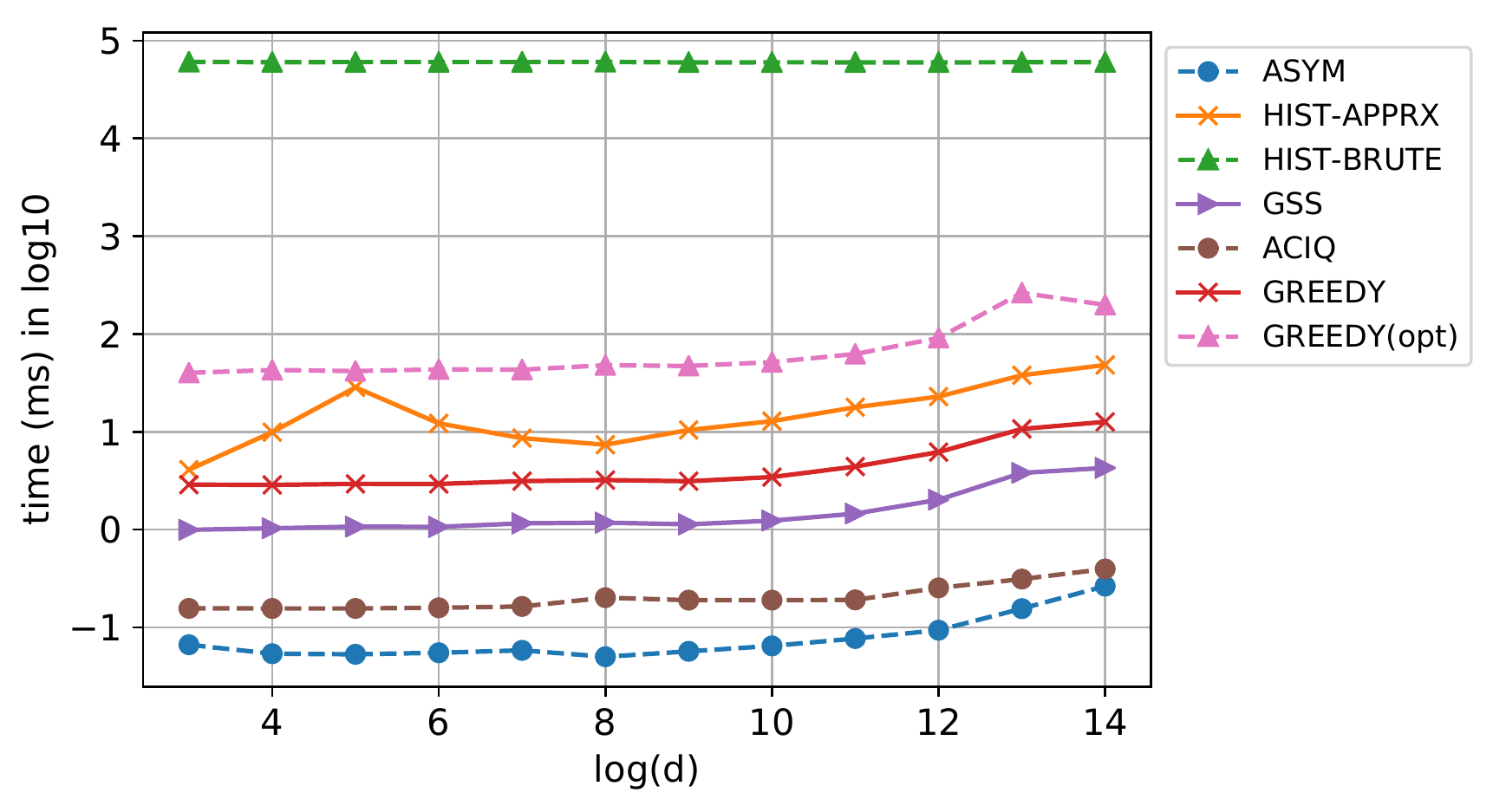}
  \end{center}
    \caption{Average time per row spent on 4-bit quantization. Time is shown in log10 scale. }
    \label{fig:time_complexity}
\end{figure}

Figure~\ref{fig:time_complexity} shows the average time in milliseconds to quantize a row vector with different dimensions using 4 bits. To make a fair comparison, we implemented all the quantization algorithms in python. The results show that HIST-BRUTE is millions of times slower than ASYM. All the other clipping-based approaches take less than 100ms to quantize a row vector when $d $ is less than $2048$.  The times are measured on a computer with Ubuntu 16.04, 3.00GHz Intel Xeon CPU E5-1607 processor, and 8GB memory. 

\section{Histograms After 4-bit Quantization}
\begin{figure}
\begin{subfigure}{.5\textwidth}
  \centering
  \includegraphics[width=.8\linewidth]{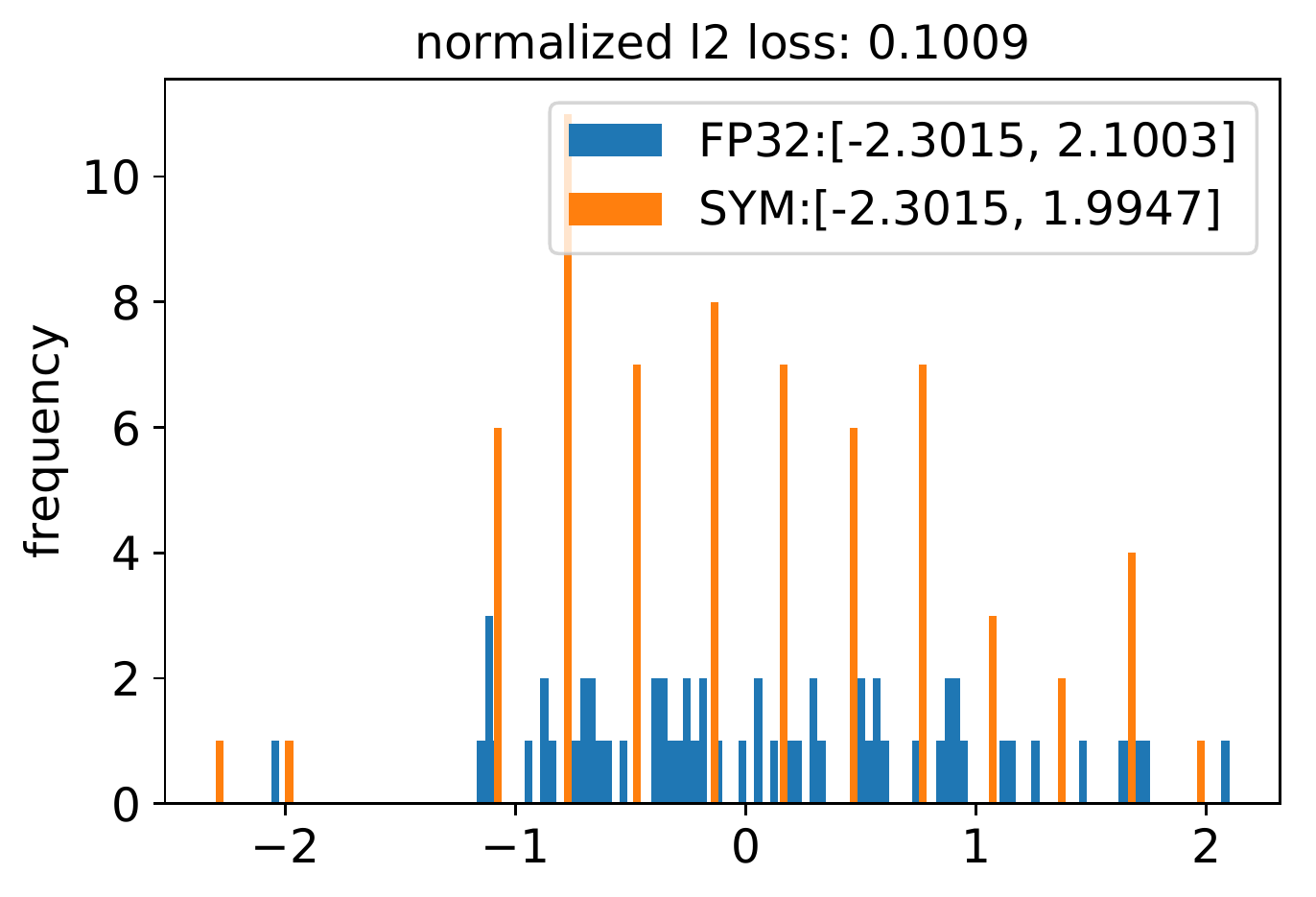}
  \caption{SYM}
  \label{fig:sym}
\end{subfigure}%
\begin{subfigure}{.5\textwidth}
  \centering
  \includegraphics[width=.8\linewidth]{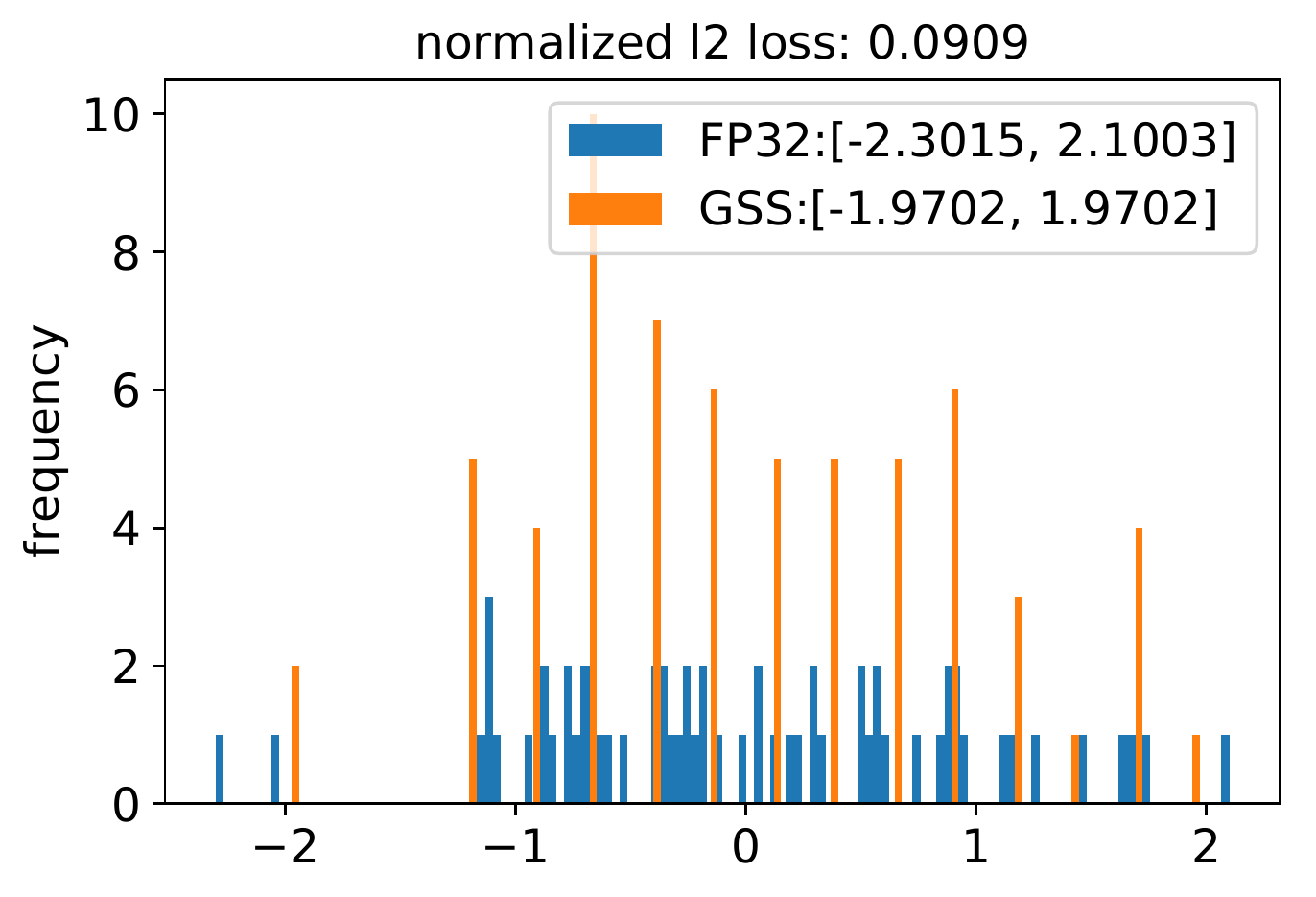}
  \caption{GSS}
  \label{fig:gss}
\end{subfigure}
\begin{subfigure}{.5\textwidth}
  \centering
  \includegraphics[width=.8\linewidth]{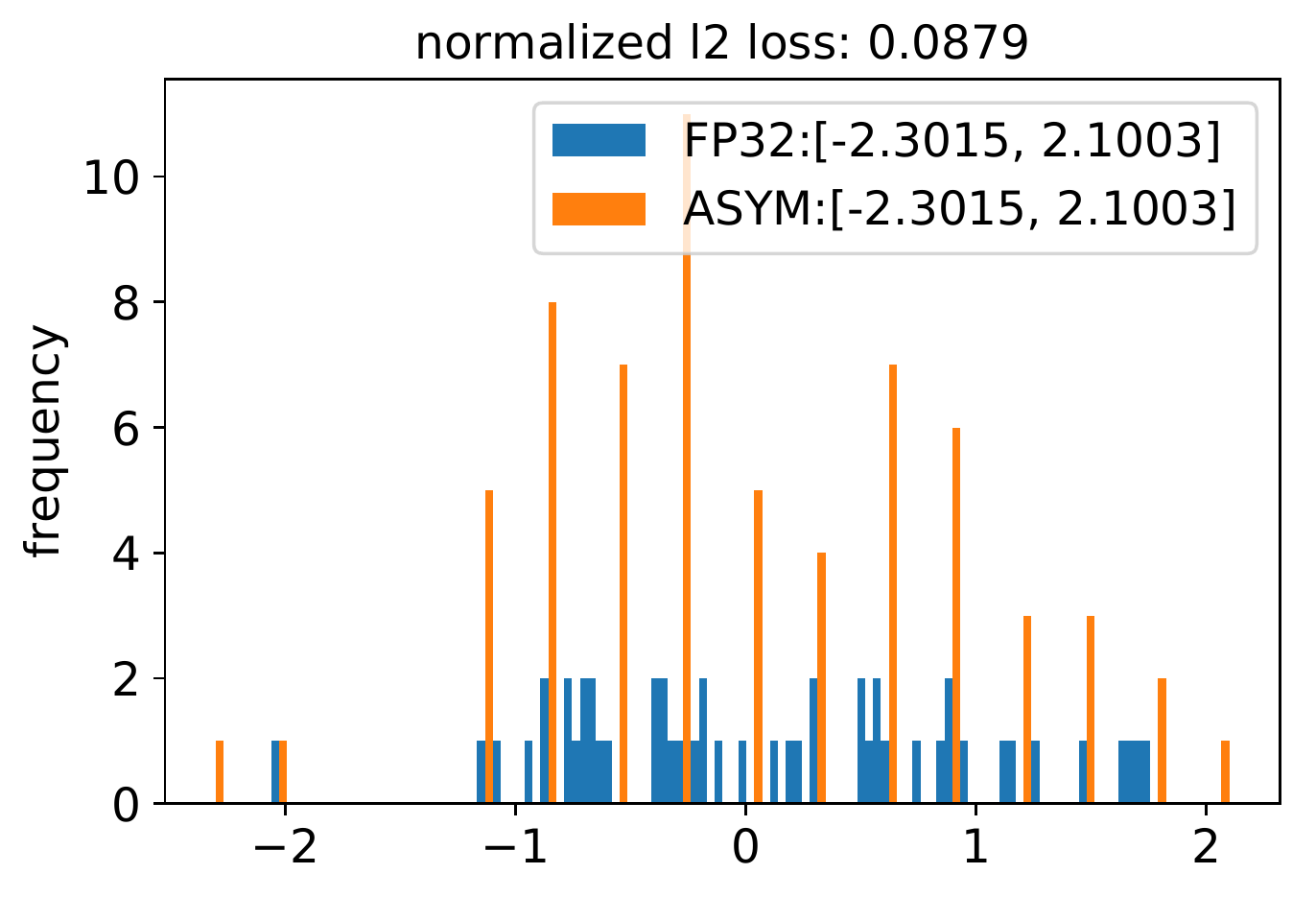}
  \caption{ASYM}
  \label{fig:asym}
\end{subfigure}%
\begin{subfigure}{.5\textwidth}
  \centering
  \includegraphics[width=.8\linewidth]{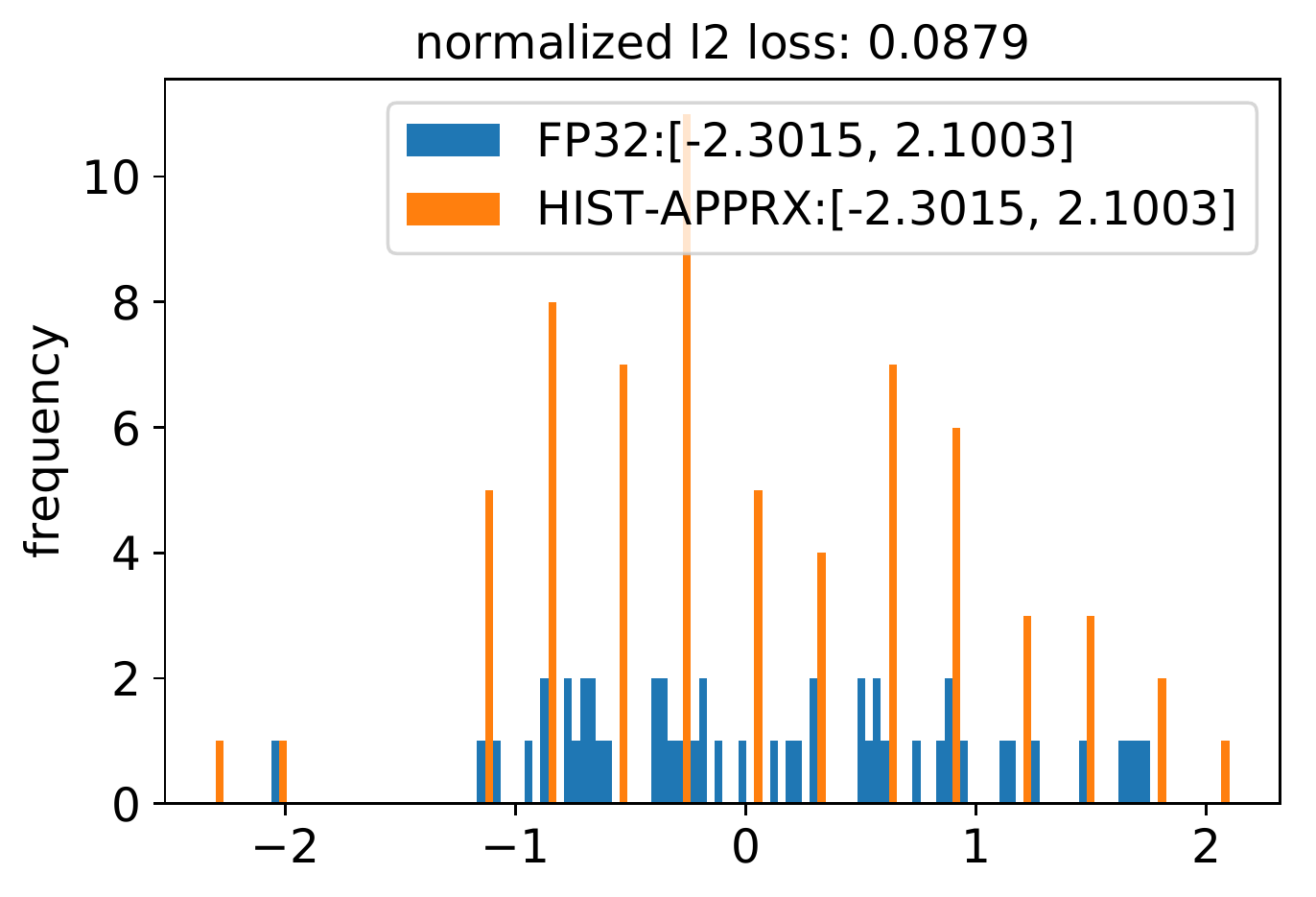}
  \caption{HIST-APPRX}
  \label{fig:approx}
\end{subfigure}
\begin{subfigure}{.5\textwidth}
  \centering
  \includegraphics[width=.8\linewidth]{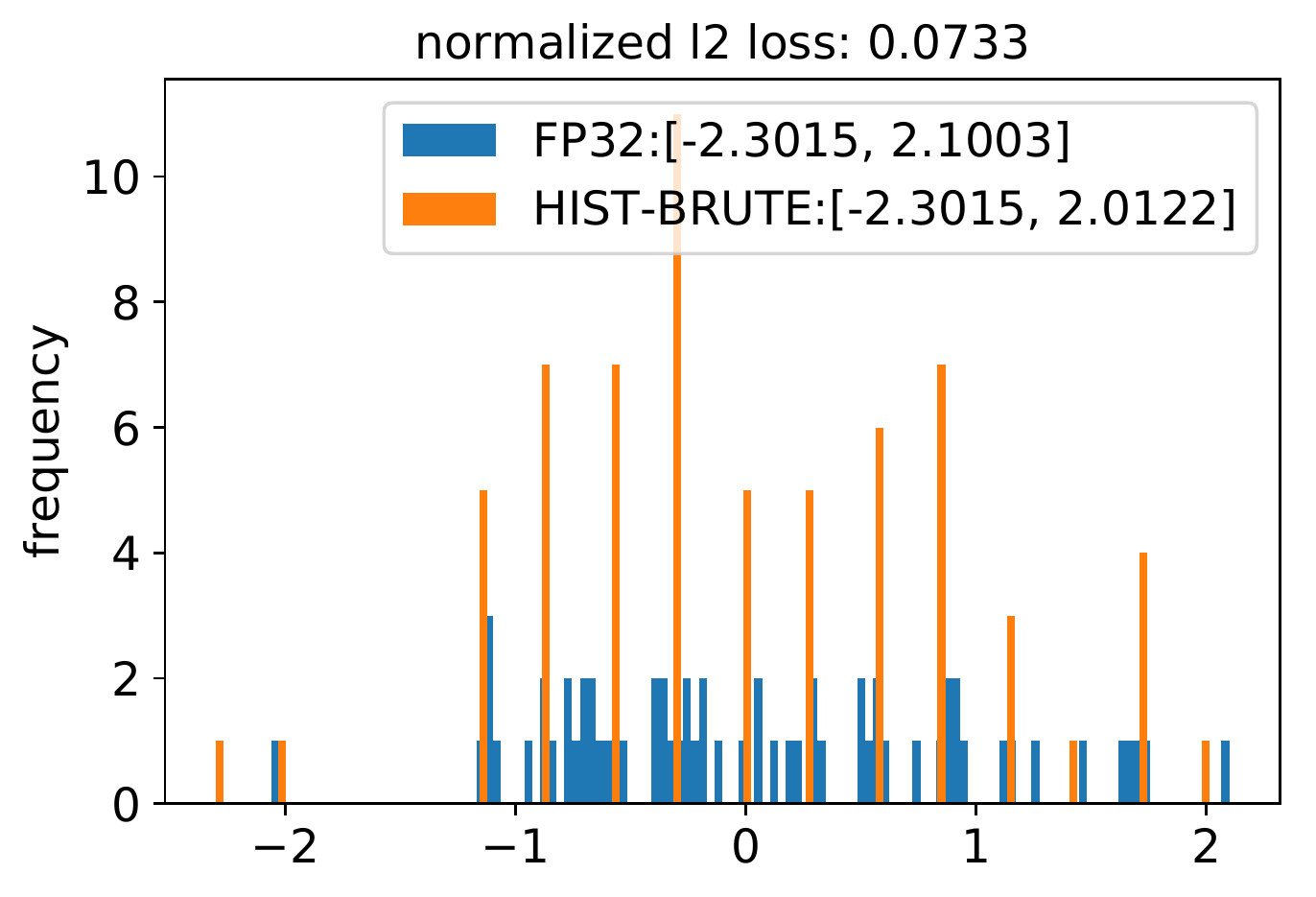}
  \caption{HIST-BRUTE}
  \label{fig:brute}
\end{subfigure}%
\begin{subfigure}{.5\textwidth}
  \centering
  \includegraphics[width=.8\linewidth]{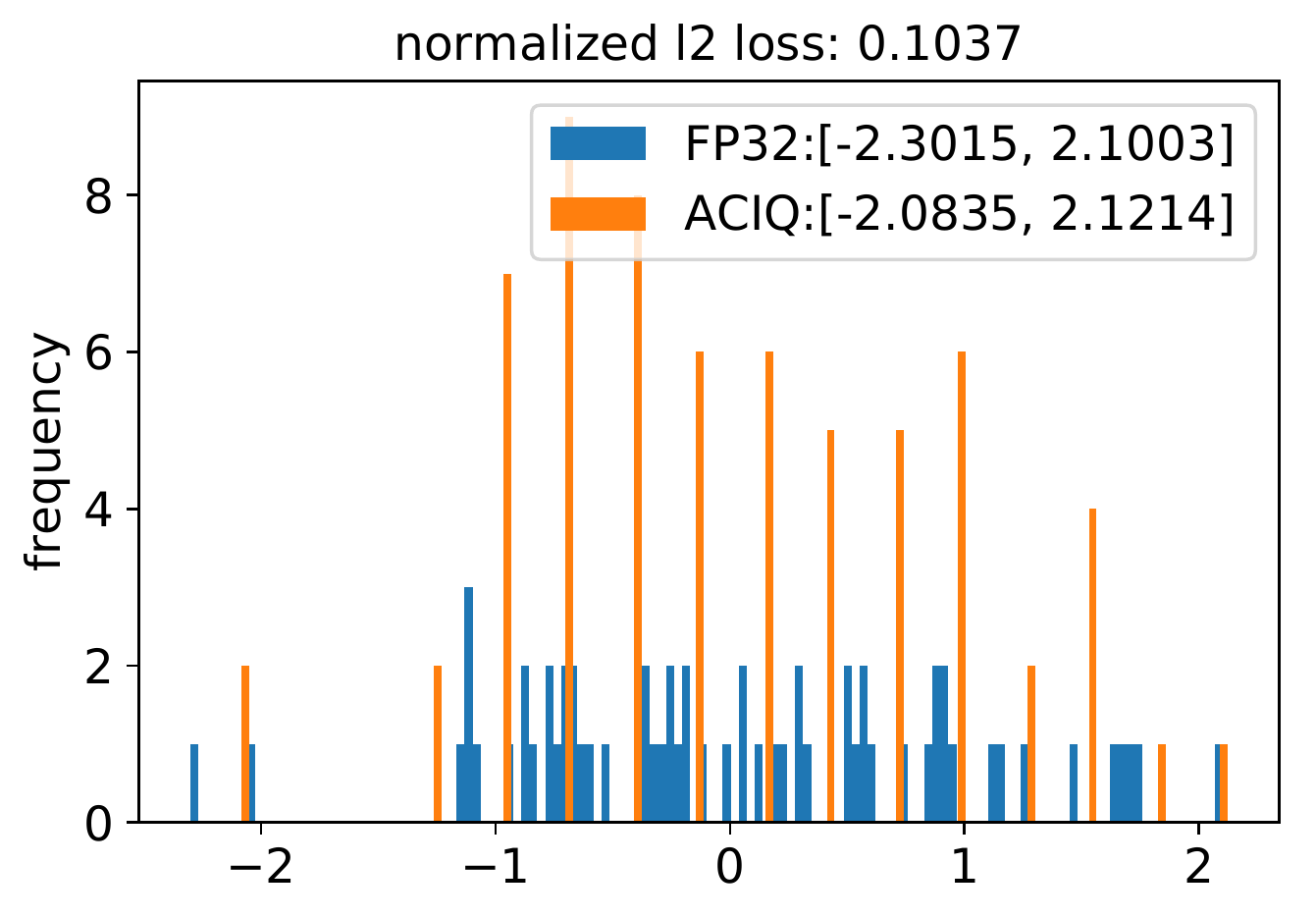}
  \caption{ACIQ}
  \label{fig:aciq}
\end{subfigure}
\begin{subfigure}{.5\textwidth}
  \centering
  \includegraphics[width=.8\linewidth]{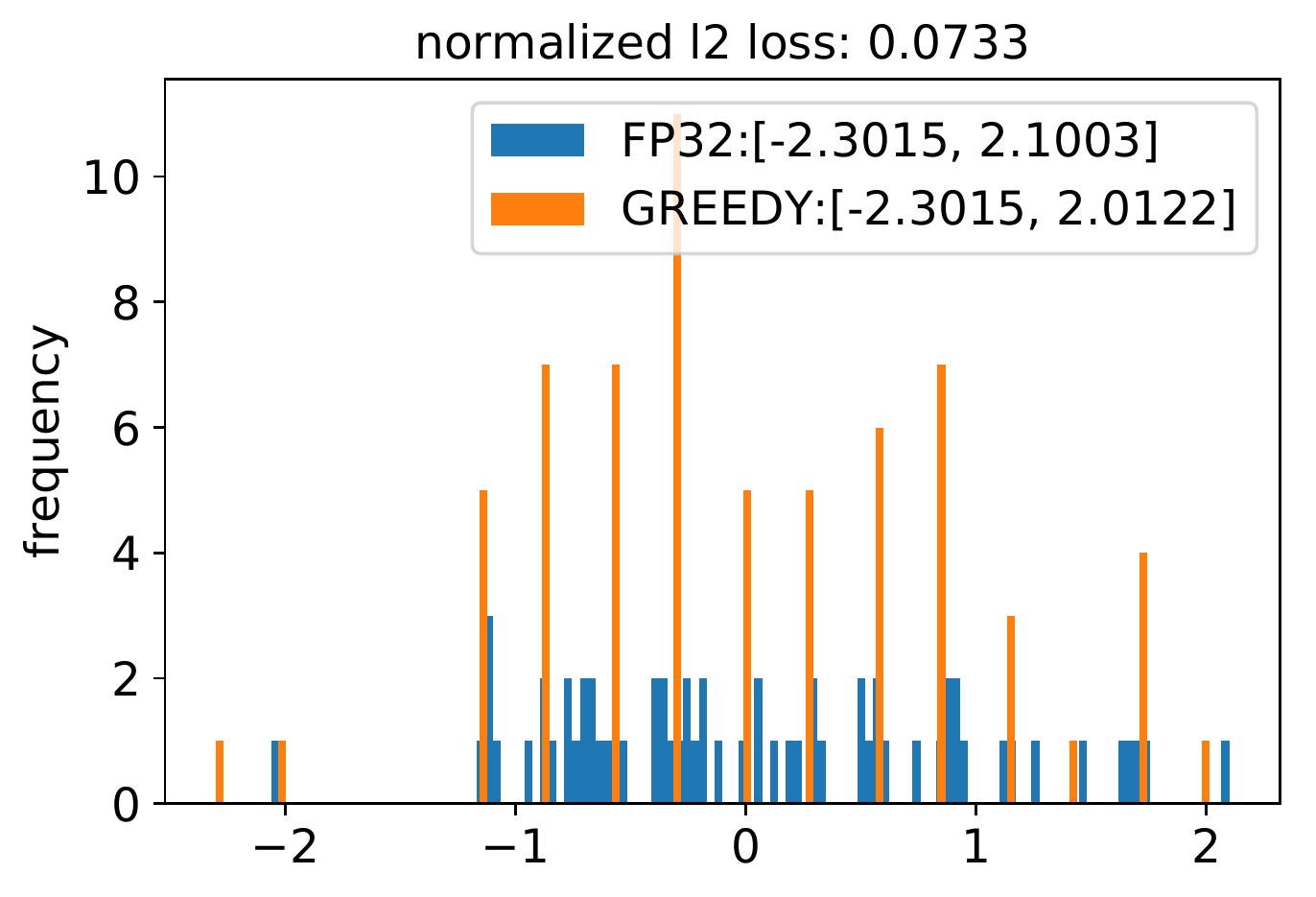}
  \caption{GREEDY}
  \label{fig:greedy}
\end{subfigure}
\begin{subfigure}{.5\textwidth}
  \centering
  \includegraphics[width=.8\linewidth]{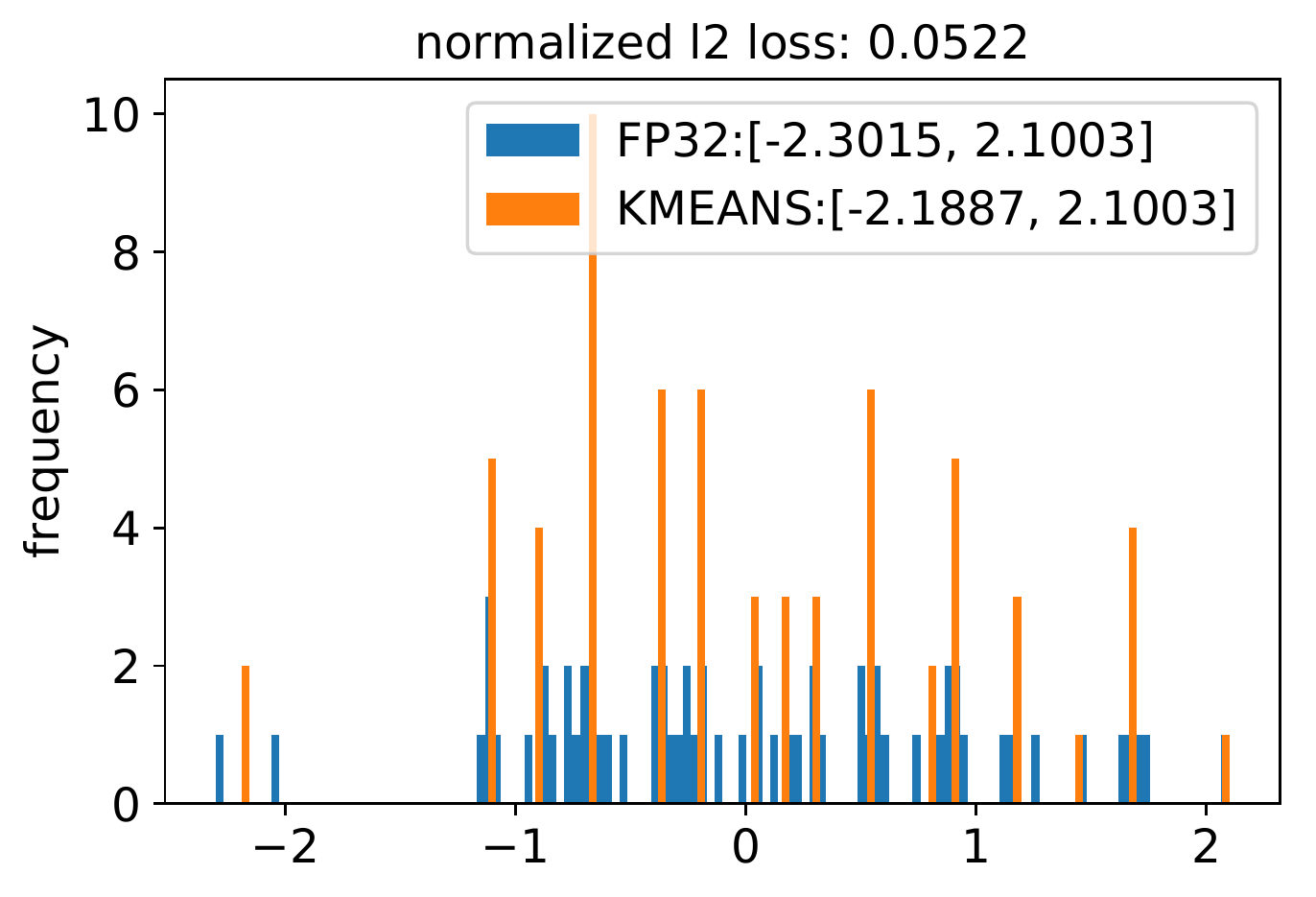}
  \caption{KMEANS}
  \label{fig:kmeans}
\end{subfigure}
\caption{Histograms of a vector (d=64) before and after 4-bit quantization with different techniques. HIST-APPRX and HIST-BRUTE use $b=200$, GREEDY uses $b=200, r=0.16$.}
\label{fig:row_vector_histogram}
\end{figure}

We show the histograms of a vector after 4-bit quantization using different approaches in Figure~\ref{fig:row_vector_histogram}. The vector is of dimension 64 and its values are randomly sampled from a Gaussian distribution.  The results echo the observations in Figure~\ref{fig:mse_diff_d} that GREEDY and KMEANS have the smallest quantization error than state-of-the-art quantization approaches (HIST-APPRX, ACIQ, GSS).

\end{document}